\documentclass{article}

\PassOptionsToPackage{round}{natbib}

\usepackage[final]{nips_2018}

\usepackage[utf8]{inputenc} 
\usepackage[T1]{fontenc}    
\usepackage{hyperref}       
\usepackage{url}            
\usepackage{booktabs}       
\usepackage{amsfonts}       
\usepackage{nicefrac}       
\usepackage{microtype}      

\usepackage{float}
\usepackage{color}
\usepackage{amssymb}
\usepackage{graphicx}
\usepackage{amsmath}
\usepackage{subfig}
\usepackage{graphicx}
\usepackage{caption}
\usepackage{bm}
\usepackage{color}
\usepackage{url}
\usepackage{appendix}
\usepackage{pdfpages}
\usepackage{bigints}
\usepackage{hyperref}

\long\def\cut#1{}

\newcommand{\SI}{{Supplementary Information}}
\newcommand{\gppvae}{GPPVAE}
\newcommand{\joint}{\gppvae-joint}
\newcommand{\dis}{\gppvae-dis}
\newcommand{\eg}{\textit{e.\,g., }}

\newcommand{\ie}{\textit{i.\,e., }}

\newcommand{\etal}{\textit{et al.}}

\newcommand{\be}{\begin{equation}}
\newcommand{\ee}{\end{equation}}
\newcommand{\bea}{\begin{eqnarray}}
\newcommand{\eea}{\end{eqnarray}}
\newcommand{\beas}{\begin{eqnarray*}}
\newcommand{\eeas}{\end{eqnarray*}}
\newcommand{\logdet}[1]{\log\left\lvert{#1}\right\rvert}

\newcommand{\iid}{{\it iid~}}

\newcommand{\given}{\,|\,}

\newcommand{\tr}[1]{\text{tr}\left(#1\right)}

\newcommand{\B}[1]{\bm{#1}}

\title{Gaussian Process Prior Variational Autoencoders}

\author{
  Francesco Paolo Casale$^{\dagger\ast}$,
  Adrian V Dalca$^{\ddagger\S}$,
  Luca Saglietti$^{\dagger\P}$,\\
  \textbf{Jennifer Listgarten$^{\sharp}$,
  Nicolo Fusi$^{\dagger}$}
}

\begin{document}

\maketitle

\vspace{-0.7cm}
\begin{centering}
$^\dagger$ Microsoft Research New England, Cambridge (MA), USA
\\
$^\ddagger$ Computer Science and Artificial Intelligence Lab, MIT, Cambridge (MA), USA
\\
$^\S$ Martinos Center for Biomedical Imaging, MGH, HMS, Boston (MA), USA;
\\
$^\P$ Italian Institute for Genomic Medicine, Torino, Italy
\\
$^\sharp$ EECS Department, University of California, Berkeley (CA), USA.
\\
\vspace{0.1cm}
$^\ast$ \texttt{frcasale@microsoft.com}
\\
\vspace{0.5cm}
\end{centering}

\begin{abstract}
Variational autoencoders (VAE) are a powerful and widely-used class of models to learn complex data distributions in an unsupervised fashion.
One important limitation of VAEs is the prior assumption that latent sample representations are independent and identically distributed.
However, for many important datasets, such as time-series of images, this assumption is too strong: accounting for covariances between samples, such as those in time, can yield to a more appropriate model specification and improve performance in downstream tasks. In this work, we introduce a new model, the Gaussian Process (GP) Prior Variational Autoencoder (\gppvae), to specifically address this issue.
The \gppvae~ aims to combine the power of VAEs with the ability to model correlations afforded by GP priors.
To achieve efficient inference in this new class of models, we leverage structure in the covariance matrix, and introduce a new stochastic backpropagation strategy that allows for computing stochastic gradients in a distributed and low-memory fashion.
We show that our method outperforms conditional VAEs (CVAEs) and an adaptation of standard VAEs in two image data applications.
\end{abstract}

\section{Introduction}
Dimensionality reduction is a fundamental approach to compression of complex, large-scale
data sets, either for visualization or for pre-processing before application of supervised approaches.
\cut{Auto-encoders represent a rich class of models to perform non-linear dimensionality reduction~\citep{vincent2008extracting}.}
Historically, dimensionality reduction has been framed in one of two modeling camps: the simple and rich capacity language of neural networks; or the probabilistic formalism of generative models, which enables Bayesian capacity control and provides uncertainty over latent encodings.
Recently, these two formulations have been combined through the Variational Autoencoder (VAE)~\citep{kingma2013auto}, wherein the expressiveness of neural networks was used to model both the mean and the variance of a simple likelihood. 
In these models, latent encodings are assumed to be identically and independently distributed (\iid) across both latent dimensions and samples.
Despite this simple prior, the model lacks conjugacy, exact inference is intractable and variational inference is used.
In fact, the main contribution of the Kingma \etal~paper is to introduce an improved, general approach for variational inference (also developed in~\cite{rezende2014stochastic}).

One important limitation of the VAE model is the prior assumption that latent representations of samples are \iid, whereas in many important problems,
accounting for sample structure is crucial for correct model specification and consequently, for optimal results. For example, in autonomous driving, or medical imaging~\citep{dalca2015predictive,lonsdale2013genotype}, high dimensional images are correlated in time---an \iid prior for these would not be sensible because, {\it a priori}, two images that were taken closer in time should have more similar latent representations than images taken further apart.
More generally, one can have multiple sequences of images from different cars, or medical image sequences from multiple patients.
Therefore, the VAE prior should be able to capture multiple levels of correlations at once, including time, object identities, {\it etc.}
A natural solution to this problem is to replace the VAE \iid prior over the latent space with a Gaussian Process (GP) prior~\citep{rasmussen2004gaussian}, which enables the specification of sample correlations through a kernel function~\citep{durrande2011additive, gonen2011multiple, wilson2013gaussian, wilson2016deep, rakitsch2013all, bonilla2007kernel}.
GPs are often amenable to exact inference, and a large body of work in making computationally challenging GP-based models tractable can be leveraged (GPs naively scale cubically in the number of samples)~\citep{gal2014distributed, bauer2016understanding, hensman2013gaussian, csato2002sparse, quinonero2005unifying, titsias2009variational}.

In this work,  we introduce the Gaussian Process Prior Variational Autoencoder (\gppvae), an extension of the VAE latent variable model where correlation between samples is modeled through a GP prior on the latent encodings. 
The introduction of the GP prior, however, introduces two main computational challenges.
First, naive computations with the GP prior have cubic complexity in the number of samples, which is impractical in most applications.
To mitigate this problem one can leverage several tactics commonly used in the GP literature, including the use of pseudo-inputs~\citep{csato2002sparse, gal2014distributed, hensman2013gaussian, quinonero2005unifying, titsias2009variational}, Kronecker-factorized covariances~\citep{casale2017joint, stegle2011efficient, rakitsch2013all}, and low rank structures \citep{casale2015efficient, lawrence2005probabilistic}.
Specifically, in the instantiations of \gppvae~considered in this paper, we focus on low-rank factorizations of the covariance matrix.
A second challenge is that the \iid assumption which guarantees unbiasedness of mini-batch gradient estimates (used to train standard VAEs) no longer holds due to the GP prior. Thus mini-batch GD is no longer applicable.
However, for the  applications we are interested in, comprising sequences of large-scale images, it is critical from a practical standpoint to avoid processing all samples simultaneously; we require a procedure that is both low in memory use and yields fast inference.
Thus, we propose a new scheme for gradient descent that enables Monte Carlo gradient estimates in a distributable and memory-efficient fashion.
This is achieved by exploiting the fact that sample correlations are only modeled in the latent (low-dimensional) space, whereas high-dimensional representations are independent when conditioning on the latent ones.

In the next sections we (i) discuss our model in the context of related work, (ii) formally develop the model and the associated inference procedure, (iii) compare \gppvae~with alternative models in empirical settings, demonstrating the advantages of our approach.

\section{Related work}

Our method is related to several extensions of the standard VAE that aim at improving the latent representation by leveraging auxiliary data, such as time annotations, pose information or lighting. 
An \textit{ad hoc} attempt to induce structure on the latent space  by grouping samples with specific properties in mini-batches was introduced in~\cite{kulkarni2015deep}.
More principled approaches proposed a semi-supervised model using a continuous-discrete mixture model that concatenates the input with auxiliary information~\citep{kingma2014semi}.
Similarly, the conditional VAE~\citep{sohn2015learning} incorporates auxiliary information in both the encoder and the decoder, and has been used successfully for sample generation with specific categorical attributes.
Building on this approach, several models use the auxiliary information in an 
unconditional way~\citep{suzuki2016joint,pandey2017variational,vedantam2017generative,wang2016deep,wu2018multimodal}.

A separate body of related work aims at designing a more flexible variational posterior distributions, either by considering a dependence on auxiliary variables~\citep{maaloe2016auxiliary}, by allowing structured encoder models~\citep{siddharth2016inducing}, or by considering chains of invertible transformations that can produce arbitrarily complex posteriors~\citep{kingma2016improved,nalisnick2016approximate,rezende2015variational}.
In other work, a dependency between latent variables is induced by way of hierarchical structures at the level of the parameters of the variational family~\citep{ranganath2016hierarchical,tran2015variational}.

The extensions of VAEs most related to~\gppvae~are those that move away from the assumption of an \iid Gaussian prior on the latent representations to consider richer prior distributions~\citep{jiang2016variational,shu2016stochastic,tomczak2017vae}.
These build on the observation that overly-simple priors can induce excessive regularization, limiting the success of such models~\citep{chen2016variational,hoffman2016elbo,siddharth2017learning}. 
For example, Johnson \etal~ proposed composing latent graphical models with deep observational likelihoods. Within their framework, more flexible priors over latent encodings are designed based on conditional independence assumptions, and a conditional random field variational family is used to enable efficient inference by way of message-passing algorithms~\citep{johnson2016composing}.

In contrast to existing methods, we propose to model the relationship between the latent space and the auxiliary information using a GP prior, leaving the encoder and decoder as in a standard VAE (independent of the auxiliary information). 
Importantly, the proposed approach allows for modeling arbitrarily complex sample structure in the data.
In this work, we specifically focus on disentangling sample correlations induced by different aspects of the data.
Additionally, \gppvae~enables estimation of latent auxiliary information when such information is unobserved by leveraging previous work~\citep{lawrence2005probabilistic}.
Finally, using the encoder and decoder networks together with the GP predictive posterior, our model provides a natural framework for out-of-sample predictions of high-dimensional data, for virtually any configuration of the auxiliary data.

\section{Gaussian Process Prior Variational Autoencoder}
\label{sec:methods}

Assume we are given a set of samples (\eg images), each coupled with different types of auxiliary data (\eg time, lighting, pose, person identity).
In this work, we focus on the case of two types of auxiliary data: \textit{object} and \textit{view} entities.
Specifically, we consider datasets with images of \textit{objects} in different \textit{views}. For example, images of faces in different poses or images of  hand-written digits at different rotation angles.
In these problems, we know both which object (person or hand-written digit) is represented in each image in the dataset, and in which view (pose or rotation angle).
Finally, each unique object and view is attached to a feature vector, which we refer to as an \textit{object feature vector} and a \textit{view feature vector}, respectively.
In the face dataset example, object feature vectors might contain face features such as skin color or hair style, while view feature vectors may contain pose features such as polar and azimuthal angles with respect to a reference position.
Importantly, as described and shown below, we can learn these feature vectors if not observed.

\subsection{Formal description of the model}

Let $N$ denote the number of samples, $P$ the number of unique objects and $Q$ the number of unique views.
Additionally, let $\{\B{y}_n\}_{n=1}^N$ denote $K$-dimensional representation for $N$ samples; let $\{\B{x}_p\}_{p=1}^P$ denote $M$-dimensional object feature vectors for the $P$ objects; and let $\{\B{w}_q\}_{q=1}^Q$ denote $R$-dimensional view feature vectors for the $Q$ views.
Finally, let $\{\B{z}_n\}_{n=1}^N$ denote the $L$-dimensional latent representations.
We consider the following generative process for the observed samples ({Fig 1a}):
\begin{itemize}
\item{the latent representation of object $p_n$ in view $q_n$ is generated from object feature vector $\B{x}_{p_n}$ and view feature vector $\B{w}_{q_n}$ as
\begin{equation}
\B{z}_n=f(\B{x}_{p_n}, \B{w}_{q_n})+\B{\eta}_n,\; \text{where}\;\B{\eta}_n\sim\mathcal{N}\left(\B{0}, \alpha\B{I}_L\right);
\end{equation}
}
\item{
image $\B{y}_n$ is generated from its latent representation $\B{z}_n$ as
\begin{equation}
\B{y}_n=g(\B{z}_{n})+\B{\epsilon}_n,\; \text{where}\;\B{\epsilon}_n\sim\mathcal{N}\left(\B{0}, \sigma_y^2\B{I}_K\right).
\end{equation}
}
\end{itemize}
The function $f:\mathbb{R}^M\times\mathbb{R}^R\rightarrow\mathbb{R}^L$ defines how sample latent representations can be obtained in terms of object and view feature vectors, while $g:\mathbb{R}^L\rightarrow\mathbb{R}^K$ maps latent representations to the high-dimensional sample space.

We use a Gaussian process (GP) prior on $f$, which allows us to model sample covariances in the latent space as a function of object and view feature vectors. Herein, we use a convolutional neural network for $g$, which is a natural choice for image data~\citep{lecun1995convolutional}.
The resulting marginal likelihood of the \gppvae, is
\begin{eqnarray}
p(\B{Y}\given\,\B{X},\B{W},\B{\phi},\sigma_y^2,\B{\theta},\alpha)=
\bigintsss{
p(\B{Y}\given\,\B{Z},\B{\phi},\sigma_y^2)
p\left(\B{Z}\given\B{X}, \B{W},\B{\theta},\alpha\right)
d\B{Z}},
\label{eq:lvm}
\end{eqnarray}
where $\B{Y}=\left[\B{y}_1, \dots,\B{y}_N\right]^T\in\mathbb{R}^{N\times{K}}$, $\B{Z}=\left[\B{z}_1, \dots,\B{z}_N\right]^T\in\mathbb{R}^{N\times{L}}$, $\B{W}=\left[\B{w}_1, \dots,\B{w}_Q\right]^T\in\mathbb{R}^{Q\times{R}}$, $\B{X}=\left[\B{x}_1, \dots,\B{x}_P\right]^T\in\mathbb{R}^{P\times{M}}$. Additionally, $\B{\phi}$ denotes the parameters of $g$ and $\B{\theta}$ the GP kernel parameters.

\paragraph{Gaussian Process Model.}
The GP prior defines the following multivariate normal distribution on latent representations:
\begin{eqnarray}
p\left(\B{Z}\given\B{X}, \B{W},\B{\theta},\alpha\right)=\prod_{l=1}^L\mathcal{N}\left(\B{z}^l\given\B{0}, \B{K}_{\B{\theta}}(\B{X}, \B{W}) + \alpha\B{I}_N\right),
\label{eq:GP}
\end{eqnarray}
where $\B{z}^l$ denotes the $l$-th column of $\B{Z}$.
In the setting considered in this paper, the covariance function $\B{K}_{\B{\theta}}$ is composed of a view kernel that models covariances between views, and an object kernel that models covariances between objects.
Specifically, the covariance between sample $n$ (with corresponding feature vectors $\B{x}_{p_n}$ and $\B{w}_{q_n}$) and sample $m$ (with corresponding feature vectors $\B{x}_{p_m}$ and $\B{w}_{q_m}$) is given by the factorized form~\citep{bonilla2007kernel, rakitsch2013all}:
\begin{equation}
\B{K}_{\B{\theta}}(\B{X}, \B{W})_{nm}=
\mathcal{K}_{\B{\theta}}^{\text{(view)}}(\B{w}_{q_n},\B{w}_{q_m})
\mathcal{K}_{\B{\theta}}^{\text{(object)}}(\B{x}_{p_n},\B{x}_{p_m}).
\label{eq:structcovfun}
\end{equation}

\paragraph{Observed versus unobserved feature vectors}
Our model can be used when either one, or both of the view/sample feature vectors are unobserved.
In this setting, we regard the unobserved features as latent variables and obtain a point estimate for them, similar to Gaussian process latent variable models~\citep{lawrence2005probabilistic}. We have done so in our experiments.

\begin{figure}[t!]
\begin{centering}
\includegraphics[width=.95 \textwidth]{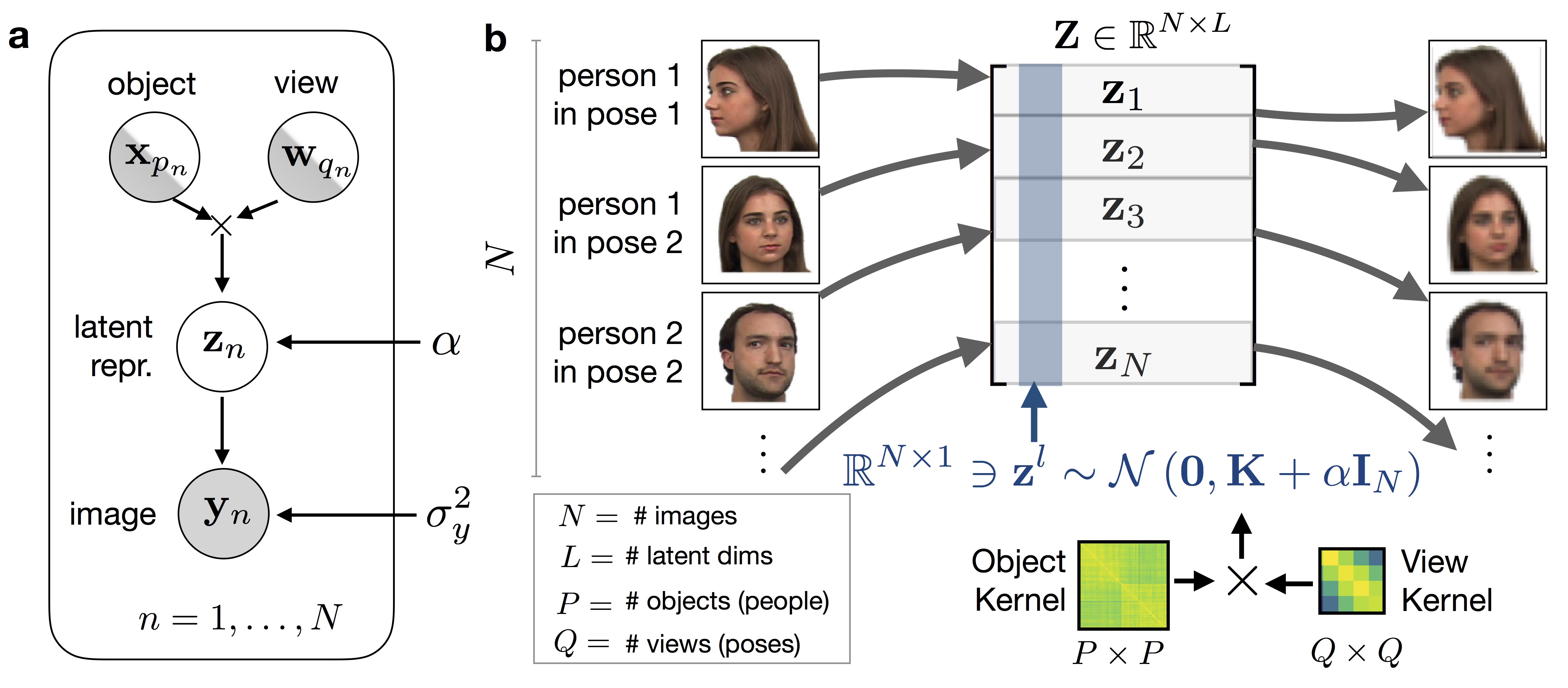}
\caption{
(\textbf{a}) Generative model underlying the proposed~\gppvae.
(\textbf{b}) Pictorial representation of the inference procedure in~\gppvae.
Each sample (here an image) is encoded in a low-dimensional space and then decoded to the original space. Covariances between samples are modeled through a GP prior on each column of the latent representation matrix $\B{Z}$.
}
\end{centering}
\end{figure}

\subsection{Inference}

As with a standard VAE, we make use of variational inference for our model.
Specifically, we consider the following variational distribution over the latent variables
\begin{eqnarray}
q_{\B{\psi}}(\B{Z}\given\B{Y})=\prod_n{\mathcal{N}\left(\B{z}_n\given\B{\mu}_{\B{\psi}}^z(\B{y}_n),\text{diag}({\B{\sigma}^z}_{\B{\psi}}^2(\B{y}_n))\right)},
\label{eq:varpost}
\end{eqnarray}
which approximates the true posterior on $\B{Z}$.
In Eq. \eqref{eq:varpost}, $\B{\mu}_{\B{\psi}}^z$ and $\B{\sigma}^z_{\B{\psi}}$ are the hyperparameters of the variational distribution and are neural network functions of the observed data, while $\B{\psi}$ denotes the weights of such neural networks.
We obtain the following evidence lower bound (ELBO):
\begin{eqnarray}
\text{log}p(\B{Y}\given\,\B{X},\B{W},\B{\phi},\sigma_y^2,\B{\theta})
&\geq&
\mathbb{E}_{\B{Z}\sim{q_{\B{\psi}}}}\left[
\sum_n
\text{log}\;\mathcal{N}(\B{y}_n\given{g_{\B{\phi}}(\B{z}_n)},\sigma^2_y\B{I}_K)
+
\text{log}p\left(\B{Z}\given\B{X}, \B{W},\B{\theta},\alpha\right)
\right]+
\nonumber
\\&&
+\frac12\sum_{nl}\text{log}({\B{\sigma}^z}_{\B{\psi}}^2(\B{y}_n)_l)+\text{const}.
\label{eq:elbp}
\end{eqnarray}

\paragraph{Stochastic backpropagation.}
We use  stochastic backpropagation to maximize the ELBO~\citep{kingma2013auto,rezende2014stochastic}.
Specifically, we approximate the expectation by sampling from a reparameterized variational posterior over the latent representations, obtaining the following loss function:
\begin{eqnarray}
l\left(\B{\phi},\B{\psi},\B{\theta},\alpha,\sigma^2_y\right)=
\;\;\;\;\;\;\;\;\;\;
\;\;\;\;\;\;\;\;\;\;
\;\;\;\;\;\;\;\;\;\;
\;\;\;\;\;\;\;\;\;\;
\;\;\;\;\;\;\;\;\;\;
\;\;\;\;\;\;\;\;\;\;
\;\;\;\;\;\;\;\;\;\;
\;\;\;\;\;\;\;\;\;\;
\;\;\;\;\;\;\;\;\;\;
\;\;\;\;\;\;\;\;\;\;
\;\;\;\;
\nonumber
\\
\;\;\;
=NK\text{log}\,\sigma^2_y
+
\underbrace{
\sum_n
\frac{
\left\lVert
\B{y}_n-g_{\B{\phi}}({\B{z}_{\B{\psi}}}_n)
\right\rVert^2
}{2\sigma_y^2}
}_{\text{reconstruction term}}
-
\underbrace{
\text{log}p\left(\B{Z}_{\B{\psi}}\given\B{X}, \B{W},\B{\theta},\alpha\right)}_{\text{latent-space GP term}}
+\underbrace{\frac12\sum_{nl}\text{log}({\B{\sigma}^z}_{\B{\psi}}^2(\B{y}_n)_l)}_{\text{regularization term}},
\label{eq:loss}
\end{eqnarray}
which we optimize with respect to $\B{\phi},\B{\psi},\B{\theta},\alpha,\sigma^2_y$.
Latent representations $\B{Z}_{\B{\psi}}=\left[{\B{z}_{\B{\psi}}}_1,\dots,{\B{z}_{\B{\psi}}}_N\right]\in\mathbb{R}^{N\times{L}}$ are sampled using the re-parameterization trick~\cite{kingma2013auto}, 
\begin{eqnarray}
{{\B{z}}_{\B{\psi}}}_n=\mu^z_{\B{\psi}}(\B{y}_n)+\B{\epsilon}_n\odot{\sigma^z}_{\B{\psi}}(\B{y}_n),\;
\B{\epsilon}_n\sim\mathcal{N}(\B{0}, \B{I}_{L\times{L}}),\;
n=1,\dots,N,
\label{eq:reptrick}
\end{eqnarray}
where $\odot$ denotes the Hadamard product.
Full details on the derivation of the loss can be found in \SI. 

\paragraph{Efficient GP computations.}
Naive computations in Gaussian processes scale cubically with the number of samples ~\citep{rasmussen2004gaussian}.
In this work, we achieve linear computations in the number of samples by assuming that the overall GP kernel is low-rank.
In order to meet this assumption, we (i) exploit that in our setting the number of views, $Q$, is much lower than the number of samples, $N$, and (ii) impose a low-rank form for the object kernel ($M\ll{N}$).
Briefly, as a result of these assumptions, the total covariance is the sum of a low-rank matrix and the identity matrix $\B{K}=\B{VV}^T + \alpha\B{I}$, where $\B{V}\in\mathbb{R}^{N\times{H}}$ and $H\ll{N}$\,~\footnote{
For example, if both the view and the object kernels are linear, we have $\B{V}=\left[\B{X}_{:,1}\odot\B{W}_{:,1}, \B{X}_{:,1}\odot\B{W}_{:,2}, \dots, \B{X}_{:,M}\odot\B{W}_{:,Q}\right]\in\mathbb{R}^{N\times{H}}$.
}.
For this covariance, computation of the inverse and the log determinant, which have cubic complexity for general covariances, can be recast to have complexity, $O(NH^2+H^3+HNK)$ and $O(NH^2+H^3)$, respectively, using the Woodbury identity~\citep{henderson1981deriving} and the determinant lemma~\citep{harville1997matrix}:
\begin{eqnarray}
\B{K}^{-1}\B{M}&=&\frac1\alpha\B{I}-\frac1\alpha\B{V}(\alpha\B{I} + \B{V}^T\B{V})^{-1}\B{V}^T\B{M},\\
\logdet{\B{K}}&=&NL\log\alpha+\logdet{\B{I} + \frac1\alpha\B{V}^T\B{V}},
\end{eqnarray}
where $\B{M}\in\mathbb{R}^{N\times{K}}$. 
Note a low-rank approximation of an arbitrary kernel can be obtained through the fully independent training conditional approximation~\citep{snelson2006sparse}, which makes the proposed inference scheme applicable in a general setting.

\paragraph{Low-memory stochastic backpropagation.}
Owing to the coupling between samples from the GP prior, mini-batch gradient descent is no longer applicable.
However, a naive implementation of full gradient descent is impractical as it requires loading the entire dataset into memory, which is infeasible with most image datasets.
To overcome this limitation, we propose a new strategy to compute gradients on the whole dataset in a low-memory fashion.
We do so by computing the first-order Taylor series expansion of the GP term of the loss with respect to both the latent encodings and the prior parameters, at each step of gradient descent. In doing so, we are able to use the following procedure:
\begin{enumerate}
\item{
Compute latent encodings from the high-dimensional data using the encoder.
This step can be performed in data mini-batches, thereby imposing only low-memory requirements.
}
\item{
Compute the coefficients of the GP-term Taylor series expansion using the latent encodings. Although this step involves computations across all samples, these have low-memory requirements as they only involve the low-dimensional representations.
}
\item{
Compute a proxy loss by replacing the GP term by its first-order Taylor series expansion, which locally has the same gradient as the original loss.
Since the Taylor series expansion is linear in the latent representations, gradients can be easily accumulated across data mini-batches, making this step also memory-efficient.}
\item{
Update the parameters using these accumulated gradients.}
\end{enumerate}
Full details on this procedure are given in \SI.

\subsection{Predictive posterior}
We derive an approximate predictive posterior for \gppvae~that enables out-of-sample predictions of high-dimensional samples.
Specifically, given training samples $\B{Y}$, object feature vectors $\B{X}$, and view feature vectors $\B{W}$, the predictive posterior for image representation $\B{y}_\star$ of object $p_{\star}$ in view $q_{\star}$ is given by
\bea
p(\B{y}_{\star}\given\B{x}_\star,\B{w}_\star,\B{Y},\B{X},\B{W})
\!\!\!&\approx&\!\!\!
\int{
\underbrace{p(\B{y}_{\star}\given\B{z}_\star)}_{\text{{\tiny decode GP prediction}}}
\underbrace{p(\B{z}_\star\given\B{x}_\star,\B{w}_\star,\B{Z},\B{X},\B{W})}_{\text{{\tiny latent-space GP predictive posterior}}}
\underbrace{q(\B{Z}\given\B{Y})}_{\text{{\tiny encode training data}}}d\B{z}_\star d\B{Z}}
\label{eq:predpost}
\eea
where $\B{x}_\star$ and $\B{w}_\star$ are object and feature vectors of object $p_{\star}$ and view $q_{\star}$ respectively, and we dropped the dependency on parameters for notational compactness.
The approximation in Eq. \eqref{eq:predpost} is obtained by replacing the exact posterior on $\B{Z}$ with the variational distribution $q(\B{Z}\given\B{Y})$ (see \SI~for full details).
From Eq. \eqref{eq:predpost}, the mean of the \gppvae~predictive posterior can be obtained by the following procedure:
(i) encode training image data in the latent space through the encoder,
(ii) predict latent representation $\B{z}_\star$ of image $\B{y}_\star$ using the GP predictive posterior, and
(iii) decode latent representation $\B{z}_\star$ to the high-dimensional image space through the decoder.

\section{Experiments}

We focus on the task of making predictions of unseen images, given specified auxiliary information.
Specifically, we want to predict the image representation of object $p$ in view $q$ when that object was never observed in that view, but assuming that object $p$ was observed in at least one other view, and that view had been observed for at least one other object.
For example, we may want to predict the pose of a person appearing in our training data set without having seen that person in that pose. To do so, we need to have observed that pose for other people. This prediction task gets at the heart of what we want our model to achieve, and therefore serves as a good evaluation metric.

\subsection{Methods considered}
In addition to the \gppvae~presented (\joint), we also considered a version with a simpler optimization scheme (\dis). We also considered two extensions of the VAE that can be used for the task at hand. Specifically, we considered:
\begin{itemize}
\item{\textbf{\gppvae~with joint optimization (\joint)}, where autoencoder and GP parameters were optimized jointly.
We found that convergence was improved by first training the encoder and the decoder through standard VAE, then optimizing the GP parameters with fixed encoder and decoder for 100 epochs, and finally, optimizing all parameters jointly.
Out-of-sample predictions from \joint~were obtained by using the predictive posterior in Eq. \eqref{eq:predpost};
}
\item{
\textbf{\gppvae~with disjoint optimization (\dis)}, where we
first learned the encoder and decoder parameters through standard VAE, and then optimized the GP parameters with fixed encoder and decoder.
Again, out-of-sample predictions were obtained by using the predictive posterior in Eq. \eqref{eq:predpost};
}
\item{
{\bf Conditional VAE (CVAE)}~\citep{sohn2015learning}, where view auxiliary information was provided as input to both the encoder and decoder networks (Figure S1, S2).
After training, we considered the following procedure to generate an image of object $p$ in view $q$.
First, we computed latent representations of all the images of object $p$ across all the views in the training data (in this setting, CVAE latent representations are supposedly independent from the view).
Second, we averaged all the obtained latent representations to obtain a unique representation of object $p$.
Finally, we fed the latent representation of object $p$ together with out-of-sample view $q$ to the CVAE decoder.
As an alternative implementation, we also tried to consider the latent representation of a random image of object $p$ instead of averaging, but the performance was worse--these results are not included;
}
\item{
\textbf{Linear Interpolation in VAE latent space (LIVAE)}, which uses linear interpolation between observed views of an object in the latent space learned through standard VAE in order to predict unobserved views of the same object.
Specifically, denoting $\B{z}_1$ and $\B{z}_2$ as the latent representations of images of a given object in views $r_1$ and $r_2$, a prediction for the image of that same object in an intermediate view, $r_\star$, is obtained by first linearly interpolating between $\B{z}_1$ and $\B{z}_2$, and then projecting the interpolated latent representation to the high-dimensional image space.
}
\end{itemize}
Consistent with the L2 reconstruction error appearing in the loss off all the aforementioned VAEs (\eg~Eq. \eqref{eq:loss}), we considered pixel-wise mean squared error (MSE) as the evaluation metric.
We used the same architecture for encoder and decoder neural networks in all methods compared (see {Figure S1, S2} in \SI).
The architecture and $\sigma_y^2$, were chosen to minimize the ELBO loss for the standard VAE on a validation set ({Figure S3}, \SI).
For CVAE and LIVAE, we also considered the alternative strategy of selecting the value of $\sigma_y^2$ that maximizes out-of-sample prediction performance on the validation set (the results for these two methods are in Figure S4 and S5).
All models were trained using the Adam optimizer~\citep{kingma2014adam} with standard parameters and a learning rate of $0.001$.
When optimizing GP parameters with fixed encoder and decoder, we observed that higher learning rates led to faster convergence without any loss in performance, and thus we used a higher learning rate of $0.01$ in this setting.

\subsection{Rotated MNIST}

\paragraph{Setup.}
We considered a variation of the MNIST dataset, consisting of rotated images of hand-written "3" digits with different rotation angles.
In this setup, objects correspond to different draws of the digit "3" while views correspond to different rotation states.
View features are observed scalars, corresponding to the attached rotation angles.
Conversely, object feature vectors are unobserved and learned from data---no draw-specific features are available.

\paragraph{Dataset generation.}
We generated a dataset from 400 handwritten versions of the digit three by rotating through $Q=16$ evenly separated rotation angles in $[0, 2\pi)$, for a total of $N=6,400$ samples.
We then kept $90\%$ of the data for training and test, and the rest for validation.
From the training and test sets, we then randomly removed 25\% of the images to consider the scenario of incomplete data.
Finally, the set that we used for out-of-sample predictions (test set) was created by removing one of the views (\ie rotation angles) from the remaining images.
This procedure resulted in $4,050$ training images spanning $15$ rotation angles and $270$ test images spanning one rotation angle.

\paragraph{Autoencoder and GP model.}
We set the dimension of the latent space to $L=16$.
For encoder and decoder neural networks we considered the convolutional architecture in Figure S1.
As view kernel, we considered a periodic squared exponential kernel taking rotation angles as inputs.
As object kernel, we considered a linear kernel taking the object feature vectors as inputs.
As object feature vectors are unobserved, we learned from data---their dimensionality was set to $M=8$.
The resulting composite kernel $\B{K}$, expresses the covariance between images $n$ and $m$ in terms of the corresponding rotations angles $w_{q_n}$ and $w_{q_m}$ and object feature vectors $\B{x}_{p_n}$ and $\B{x}_{p_m}$ as
\begin{equation}
\B{K}_{\B{\theta}}(\B{X}, \B{w})_{nm}=
\beta
\underbrace{\text{exp}\left(-\frac{2\text{sin}^2{\left|w_{q_n}-w_{q_m}\right|}}{\nu^2}\right)}_{\text{rotation kernel}}
\cdot
\underbrace{\B{x}_{p_n}^T\B{x}_{p_m}}_{\text{digit draw kernel}},
\end{equation}
where $\beta\geq0$ and $\nu\geq0$ are kernel hyper-parameters learned during training of the model~\citep{rasmussen2004gaussian}, and we set $\B{\theta}=\{\beta, \nu\}$.

\paragraph{Results.}
\joint~ and \dis~ yielded lower MSE than CVAE and LIVAE in the interpolation task, with \joint~performing significantly better than \dis~($0.0280\pm0.0008$ for GPVAE-joint vs $0.0306\pm0.0009$ for \dis, $p<0.02$, \textbf{Fig. 2a},\textbf{b}).
Importantly, \joint~ learns different variational parameters than a standard VAE ({Fig. 2c,d}), used also by \dis, consistent with the fact that \joint~ performs better by adapting the VAE latent space using guidance from the prior.

\begin{figure}[!t]
\begin{centering}
\includegraphics[width=.99 \textwidth]{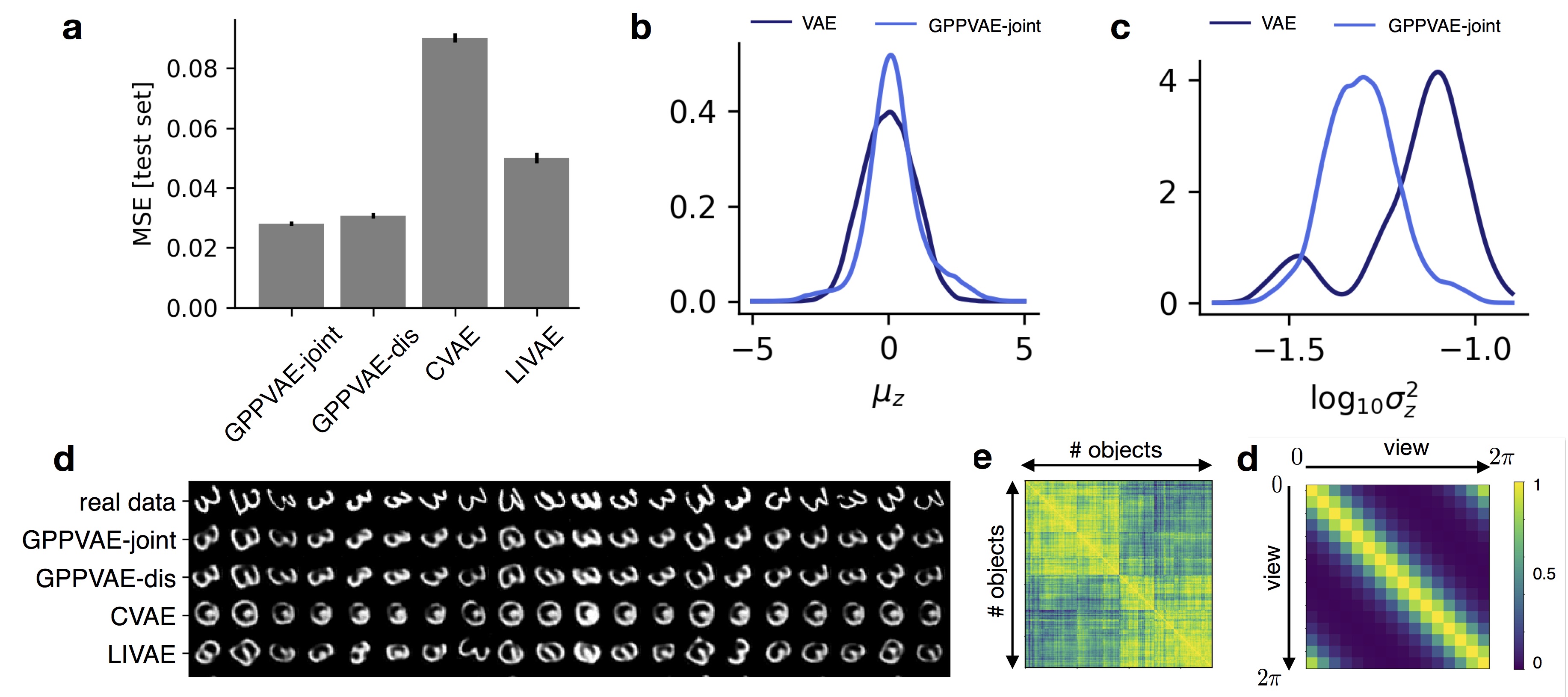}
\caption{
Results from experiments on rotated MNIST.
(\textbf{a})
Mean squared error on test set.
Error bars represent standard error of per-sample MSE.
(\textbf{b})
Empirical density of estimated means of $q_{\B{\psi}}$, aggregated over all latent dimensions.
(\textbf{c})
Empirical density of estimated log variances of $q_{\B{\psi}}$.
(\textbf{d})
Out-of-sample predictions for ten random draws of digit "3" at the out-of-sample rotation state.
(\textbf{e}, \textbf{f})
Object and view covariances learned through \joint.
}
\end{centering}
\end{figure}

\subsection{Face dataset}

\paragraph{Setup.}
As second application, we considered the Face-Place  Database (3.0)~\citep{righi2012recognizing}, which contains images of people faces in different poses.
In this setting, objects correspond to the person identities while views correspond to different poses.
Both view and object feature vectors are unobserved and learned from data.
The task is to predict images of people face in orientations that remained unobserved.

\paragraph{Data.}
We considered 4,835 images from the Face-Place  Database (3.0), which includes images of faces for 542 people shown across nine different poses (frontal and 90, 60, 45, 30 degrees left and right\footnote{
We could have used the pose angles as view feature scalar similar to the application in rotated MNIST, but purposely ignored these features to consider a more challenging setting were neither object and view features are observed.
}).
We randomly selected $80\%$ of the data for training ($n=3,868$), $10\%$ for validation ($n=484$) and $10\%$ for testing ($n=483$).
All images were rescaled to ${128}\times{128}$.

\paragraph{Autoencoder and GP model.}
We set the dimension of the latent space to $L=256$.
For encoder and decoder neural networks we considered the convolutional architecture in Figure S2.
We consider a full-rank covariance as a view covariance (only nine poses are present in the dataset) and a linear covariance for the object covariance ($M=64$).

\paragraph{Results.}
\joint and \dis yielded lower MSE than CVAE and LIVAE (\textbf{Fig. 2a},\textbf{b}).
In contrast to the MNIST problem, the difference between \joint~and\dis~was not significant ($0.0281\pm0.0008$ for \joint~vs $0.0298\pm0.0008$ for \dis).
Importantly, \joint~was able to dissect people (object) and pose (view) covariances by learning people and pose kernel jointly (\textbf{Fig. 2a},\textbf{b}).

\begin{figure}[!t]
\begin{centering}
\includegraphics[width=.95 \textwidth]{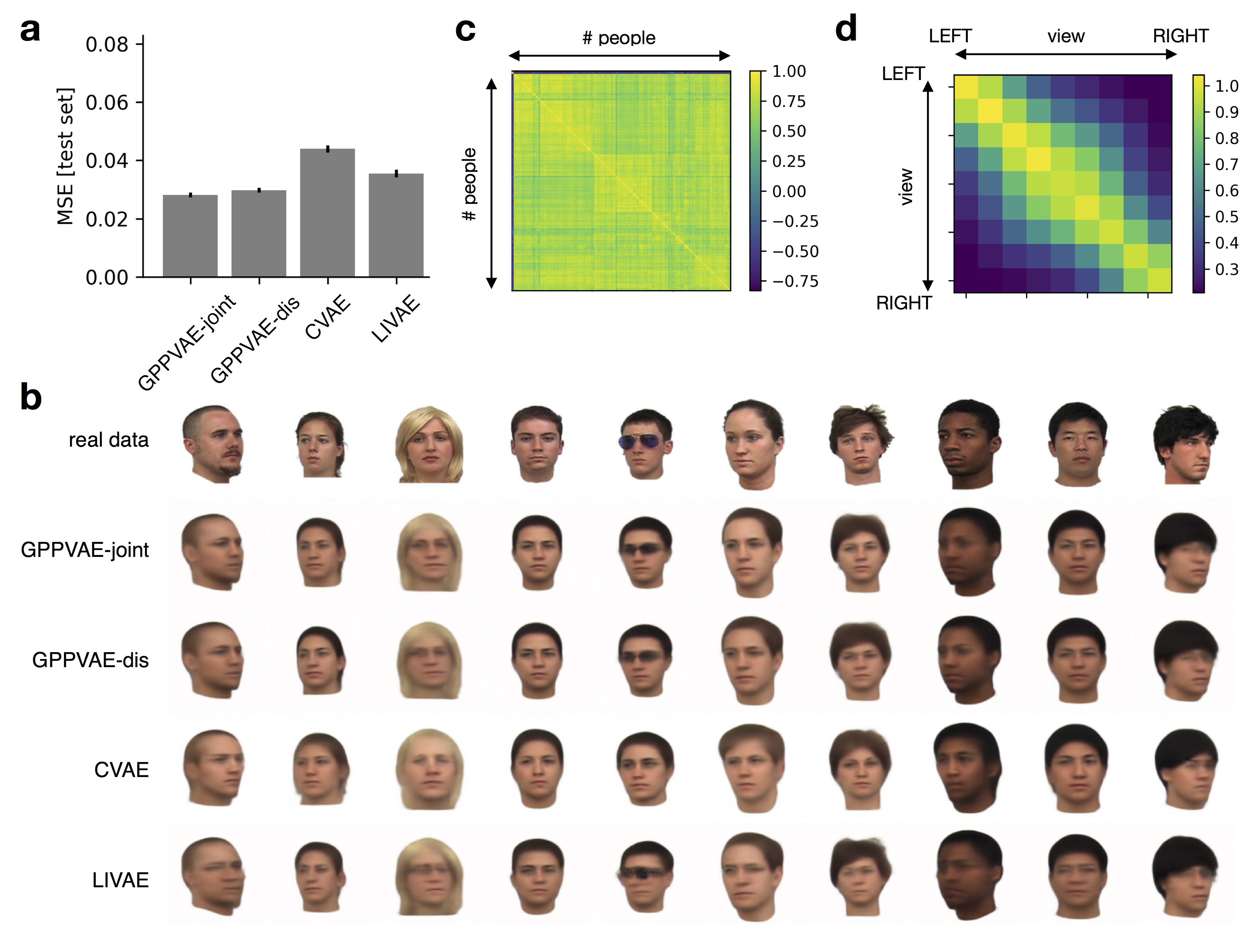}
\caption{
 Results from experiments on the face dataset.
(\textbf{a})
Mean squared error on test set 
(\textbf{b})
Out-of-sample predictions of people faces in out-of-sample poses.
(\textbf{c}, \textbf{d})
Object and view covariances learned through \joint.
}
\end{centering}
\end{figure}

\section{Discussion}
\label{sec:discussion}

We introduced \gppvae, a generative model that incorporates a GP prior over the latent space.
We also presented a low-memory and computationally efficient inference strategy for this model, which makes the model applicable to large high-dimensional datasets.
\gppvae~outperforms natural baselines (CVAE and linear interpolations in the VAE latent space) when predicting out-of-sample test images of objects in specified views (\eg pose of a face, rotation of a digit).
Possible future work includes augmenting the \gppvae~loss with a discriminator function, similar in spirit to a GAN~\citep{goodfellow2014generative}, or changing the loss to be perception-aware~\citep{hou2017deep} (see results from preliminary experiments in~{Figure S6}).
Another extension is to consider approximations of the GP likelihood that fully factorize over data points~\citep{hensman2013gaussian}; this could further improve the scalability of our method.

\subsubsection*{Code availability}
An implementation of \gppvae~is available at \url{https://github.com/fpcasale/GPPVAE}.

\subsubsection*{Acknowledgments}
Stimulus images courtesy of Michael J. Tarr, Center for the Neural Basis of Cognition and Department of Psychology, Carnegie Mellon University, http:// www.tarrlab.org. Funding provided by NSF award 0339122.

\bibliography{paper}
\bibliographystyle{plainnat}

\clearpage

\begin{appendices}

\section{Supplementary Information for Gaussian Process Prior Variational Autoencoders}

\subsection{Supplementary Methods}

\subsubsection*{Derivation of the ELBO}

We assume the following posterior distribution on latent variables
\begin{eqnarray}
q_{\B{\psi}}(\B{Z}\given\B{Y})=\prod_n{\mathcal{N}\left(\B{z}_n\given\B{\mu}_{\B{\psi}}^z(\B{y}_n),\text{diag}({\B{\sigma}^z}_{\B{\psi}}^2(\B{y}_n))\right)},
\end{eqnarray}
where $\B{\psi}$ are variational parameters. The evidence lower bound (ELBO) can be derived as follows:

\footnotesize
\begin{eqnarray}
\text{log}p(\B{Y}\given\,\B{X},\B{W},\B{\phi},\sigma_y^2,\B{\theta})
&=&
\text{log}
\int
\frac{p(\B{Y}\given\,\B{z},\B{\phi},\sigma_y^2)
p\left(\B{Z}\given\B{X}, \B{W},\B{\theta},\alpha\right)}
{q_{\B{\psi}}(\B{Z}\given\B{Y})}
q_{\B{\psi}}(\B{Z}\given\B{Y})d\B{Z}
\nonumber
\\
&\geq&
\int
\text{log}\left(\frac{p(\B{Y}\given\,\B{z},\B{\phi},\sigma_y^2)
p\left(\B{Z}\given\B{X}, \B{W},\B{\theta},\alpha\right)}
{q_{\B{\psi}}(\B{Z}\given\B{Y})}\right)
q_{\B{\psi}}(\B{Z}\given\B{Y})d\B{Z}
\nonumber
\\
&=&
\mathbb{E}_{\B{Z}\sim{q_{\B{\psi}}}}\left[
\sum_n
\text{log}\;\mathcal{N}(\B{y}_n\given{g_{\B{\phi}}(\B{z}_n)},\sigma^2_y\B{I}_K)
+
\text{log}p\left(\B{Z}\given\B{X}, \B{W},\B{\theta},\alpha\right)
\right]+
\\
&&
-
\int
\text{log}q_{\B{\psi}}(\B{Z}\given\B{Y})
q_{\B{\psi}}(\B{Z}\given\B{Y})d\B{Z}
\nonumber
\\
&=&
\mathbb{E}_{\B{Z}\sim{q_{\B{\psi}}}}\left[
\sum_n
\text{log}\;\mathcal{N}(\B{y}_n\given{g_{\B{\phi}}(\B{z}_n)},\sigma^2_y\B{I}_K)
+
\text{log}p\left(\B{Z}\given\B{X}, \B{W},\B{\theta},\alpha\right)
\right]+
\\
&&
+\frac12\sum_{nl}\text{log}({\B{\sigma}^z}_{\B{\psi}}^2(\B{y}_n)_l)+\text{const}
\nonumber
\end{eqnarray}
\normalsize

\subsubsection*{Derivation of the loss function}
First, we approximate the expectation by sampling.
Specifically, we sample a latent representation $\B{Z}_{\B{\psi}}=\left[{\B{z}_{\B{\psi}}}_1,\dots,{\B{z}_{\B{\psi}}}_N\right]\in\mathbb{R}^{N\times{L}}$  from the posterior as:
\begin{eqnarray}
{{\B{z}}_{\B{\psi}}}_n=\mu^z_{\B{\psi}}(\B{y}_n)+\B{\epsilon}_n\odot{\sigma^z}_{\B{\psi}}(\B{y}_n),\;
\B{\epsilon}_n\sim\mathcal{N}(\B{0}, \B{I}_{L\times{L}}),\;
n=1,\dots,N,
\end{eqnarray}
where we used the reparametrization trick to separate the noisy generation of samples from the model parameters.
The resulting approximate ELBO is
\footnotesize
\begin{eqnarray}
\text{ELBO}
&\approx&
\sum_n
\text{log}\;\mathcal{N}(\B{y}_n\given{g_{\B{\phi}}({\B{z}_{\B{\psi}}}_n)},\sigma^2_y\B{I}_K)
+
\text{log}p\left(\B{Z}_{\B{\psi}}\given\B{X}, \B{W},\B{\theta},\alpha\right)
+\frac12\sum_{nl}\text{log}({\B{\sigma}^z}_{\B{\psi}}^2(\B{y}_n)_l)+\text{const}
\nonumber
\end{eqnarray}
\normalsize
Finally, as we optimize $\sigma_y^2$ on a validation set (and not on the training set) in order to avoid overfitting, the maximization of the approximate ELBO on the training set is equivalent to minimizing the following cost:
\begin{eqnarray}
{\footnotesize
\text{loss}=
\frac{1}{K}\underbrace{\sum_i\left(\B{y}_i-g_{\B{\phi}}({\B{\omega}_{\B{\psi}}}_n)\right)^2}_{\text{reconstruction term}}
-
\frac{\lambda(\sigma^2_y)}{L}\left[
\underbrace{
\text{log}p\left(\B{Z}_{\B{\psi}}\given\B{X}, \B{W},\B{\theta},\alpha\right)}_{\text{latent-space GP model}}
+\underbrace{\frac12\sum_{nl}\text{log}({\B{\sigma}^z}_{\B{\psi}}^2(\B{y}_n)_l)}_{\text{regularization}}\right],
\label{eq:loss}
}
\end{eqnarray}
where $K$ is the number of pixels in the image and $\lambda$ is a trade-off parameter balancing data reconstruction and latent space prior regularization.

\paragraph{Selection of $\lambda$.}
We select the value of $\lambda$ based on a standard VAE and use the same value for all compared models.
Specifically, we optimize the VAE loss for different values of $\lambda$ and select the value for which the VAE ELBO is maximal on a validation set.
In order to compute the VAE validation ELBO we need to estimate optimal value of $\sigma^2_y$ for every value of $\lambda$ in the grid.
Given a certain value of $\lambda=\hat{\lambda}$, the optimal value of $\sigma^2_y$ can be estimated as
\begin{equation}
{\sigma^2_y}^{\text{(val)}}=\frac1{N^{\text{(val)}}}\sum_{n=1}^{N^{\text{(val)}}}\left(\B{y}^{\text{(val)}}_n-g_{\B{\phi_\lambda}}({\B{\B{z}}_{\B{\psi_{\hat{\lambda}}}}}_n^{\text{(val)}})\right)^2,
\end{equation}
where $N^{\text{(val)}}$ is the number of samples in the validation set and ($\B{\phi}_{\hat{\lambda}}$, $\B{\psi}_{\hat{\lambda}}$) are the values of the encoder/decoder parameters obtained for $\lambda=\hat{\lambda}$.

\subsubsection*{Fast low-rank computations}

Parameter inference in Gaussian models scales cubically with the number of observations.
To preform fast parameter inference in our Gaussian model, we use the fact that the total covariance is the sum of low-rank matrix and the identity matrix:
\begin{eqnarray}
\B{K}=\B{VV}^T + \alpha\B{I}
\end{eqnarray} 
where $\B{V}\in\mathbb{R}^{N\times{O}}$ and $O\ll{N}$.
Using the Woodbury identity~\cite{henderson1981deriving} and the determinant lemma~\cite{harville1997matrix}, the linear system $\B{K}^{-1}\B{M}$ with $M\in\mathbb{R}^{N\times{K}}$ and the log determinant of $\B{K}$ can be computed as:
\begin{eqnarray}
\B{K}^{-1}\B{M}&=&\frac1\alpha\B{I}-\frac1\alpha\B{V}(\alpha\B{I} + \B{V}^T\B{V})^{-1}\B{V}^T\B{M},\\
\text{logdet}\B{K}&=&NL\text{logdet}\alpha+\text{logdet}(\B{I} + \frac1\alpha\B{V}^T\B{V}),
\end{eqnarray}
which have $O(NO^2+O^3+ONK)$ and $O(NO^2+O^3)$ complexities, respectively.

\subsubsection*{Implementation of low-memory stochastic backpropagation}

For a single latent dimension the GP prior introduces the following term in the ELBO:
\begin{eqnarray}
\text{log}p\left(\B{z}_{\B{\psi}}\given\B{X}, \B{W},\B{\theta},\alpha\right)=-\frac12\B{z}_{\B{\psi}}^T\B{K}_{\theta}(\B{X}, \B{V})^{-1}\B{z}_{\B{\psi}}-\frac12\text{logdet}\B{K}_{\theta}(\B{X}, \B{V})
\label{eq:gpprior1dim}
\end{eqnarray}
In the equation above, $\B{z}_{\B{\psi}}$ functionally depends on the image space representations and thus a naive computation of full gradient descent would require loading the entire high-dimensional dataset in memory, which would be unfeasible form many high-dimensional image datasets.
To overcome this limitation, we recast computations in a form where full-matrix operations only take place in the low dimensional space, while the dependency on the nested derivatives, which involve high memory loads, is linearized and mini-batch operations are therefore allowed.
This can be achieved by considering a first-order Taylor expansion of the GP prior term.
Specfically, using that $\B{K}_{\B{\theta}}=\B{V}_{\B{\theta}}\B{V}_{\B{\theta}}^T + \alpha\B{I}$, and collecting all parameters in $\B{\xi}=\{\B{\psi},\B{\theta},\alpha\}$, we can rewrite the the term in Eq. $\eqref{eq:gpprior1dim}$ in this functional form $f(\B{z}(\B{\xi}), \B{V}(\B{\xi}), \alpha(\B{\xi}))$.
The first-order Taylor expansion of $f(\B{z}(\B{\xi}), \B{V}(\B{\xi}), \alpha(\B{\xi}))$ around $(\B{z}_{\B{\xi}_0}, \B{V}_{\B{\xi}_0}, \alpha_{\B{\xi}_0})$ is:
\begin{eqnarray}
f(\B{z}(\B{\xi}), \B{V}(\B{\xi}), \alpha(\B{\xi})) \approx \B{a}^T\B{z}(\B{\xi})+\tr{\B{B}^T\B{V}(\B{\xi})}+ c\alpha(\B{\xi})+\text{const}.
\end{eqnarray}
where
\bea
\B{a} &=& \left(\frac{\partial{f}}{\partial{\B{z}}}\right)_{\B{\xi}_0}=
\left(\B{K}^{-1}\B{z}\right)_{\B{\xi}_0}\\
\B{B} &=& \left(\frac{\partial{f}}{\partial{\B{V}}}\right)_0=
\left(-\B{K}^{-1}\B{z}\B{z}^TK^{-1}\B{V}+\B{K}^{-1}\B{V}\right)_{\B{\xi}_0}\\
c &=& \left(\frac{\partial{f}}{\partial{\alpha}}\right)_{\B{\xi}_0}=
\frac12\left(-\B{z}^T\B{K}^{-1}\B{K}^{-1}\B{z} + \text{tr}(\B{K}^{-1})\right)_{\B{\xi}_0}
\eea
\\
Note that this approximation, applied at every step of gradient descent, locally preserves the gradients.
Thus we can engineer the following low-memory four-step procedure for full gradient descent:
\begin{itemize}
\item{produce and store latent noise realizations ($\B{\epsilon}$) used for the reparametrization trick;}
\item{obtain latent variable representations $\B{z}$, combining the outputs of the encoder and the noise $\B{\epsilon}$. We employ mini-batch forward propagation for this step;}
\item{evaluate $\B{a}$, $\B{B}$ and $c$ across all samples.
Note that this step has low-memory requirements as it only involes low-dimensional representations;}
\item{exploit the local Taylor expansion as a proxy for our optimization.
As this function is linear in the data, we can accumulate its gradient in a mini-batch fashion, by passing mini-batches of $\B{\epsilon}$, $\B{Y}$, $\B{a}$ and $\B{B}$;}
\item{update the parameters $\B{\xi}$ using the full gradients as in standard gradient descent.}
\end{itemize}
Finally, we note that in our setting, the computation of $\B{a}$, $\B{B}$ and $c$ has linear complexity in the number of samples because of the low rank structure of $\B{K}$ (see previous section).

\subsubsection*{Out-of-sample predictions.}
We here derive an approximate predictive posterior for \gppvae.
Specifically, given training images $\B{Y}$, object representations $\B{X}$ and view representations $\B{V}$ the predictive posterior for image $\B{y}_\star$ representing object $\B{x}_\star$ in view $\B{v}_\star$ is
\footnotesize
\bea
p(\B{y}_{\star}\given\B{x}_\star,\B{v}_\star,\B{Y},\B{X},\B{V})
&=&
\frac{p(\B{y}_{\star},\B{Y}\given\B{x}_\star,\B{v}_\star,\B{X},\B{V})}{p(\B{Y}\given\B{X},\B{V})}\\
&=&
\frac1{p(\B{Y}\given\B{X},\B{V})}\int{p(\B{y}_{\star}\given\B{z}_\star)p(\B{Y}\given\B{Z})\underbrace{p(\B{z}_\star, \B{Z}\given\B{x}_\star,\B{v}_\star,\B{X},\B{V})}_{\text{joint distribution}}d\B{z}_\star d\B{Z}}\\
&=&
\frac1{p(\B{Y}\given\B{X},\B{V})}\int{p(\B{y}_{\star}\given\B{z}_\star)p(\B{Y}\given\B{Z})
\underbrace{p(\B{z}_\star\given\B{x}_\star,\B{v}_\star,\B{Z},\B{X},\B{V})}_{\text{predictive posterior}}
\underbrace{p(\B{Z}\given\B{X},\B{V})}_{\text{GP prior on training}}d\B{z}_\star d\B{Z}}\\
&=&
\int{p(\B{y}_{\star}\given\B{z}_\star)
\underbrace{p(\B{z}_\star\given\B{x}_\star,\B{v}_\star,\B{Z},\B{X},\B{V})}_{\text{predictive posterior}}
\underbrace{\frac{p(\B{Y}\given\B{Z})p(\B{Z}\given\B{X},\B{V})}{p(\B{Y}\given\B{X},\B{V})}}_{\text{posterior on $\B{Z}$}}d\B{z}_\star d\B{Z}}\\
&\approx&
\int{p(\B{y}_{\star}\given\B{z}_\star)
\underbrace{p(\B{z}_\star\given\B{x}_\star,\B{v}_\star,\B{Z},\B{X},\B{V})}_{\text{predictive posterior}}
\underbrace{q(\B{Z}\given\B{Y})}_{\text{approx. post. on $\B{Z}$}}d\B{z}_\star d\B{Z}}
\eea
\normalsize
where we dropped the dependency from the model and the variational parameters to simplify the notation.

\subsubsection{Supplementary Figures}

\begin{figure}[h]
\begin{centering}
\includegraphics[width=0.5 \textwidth]{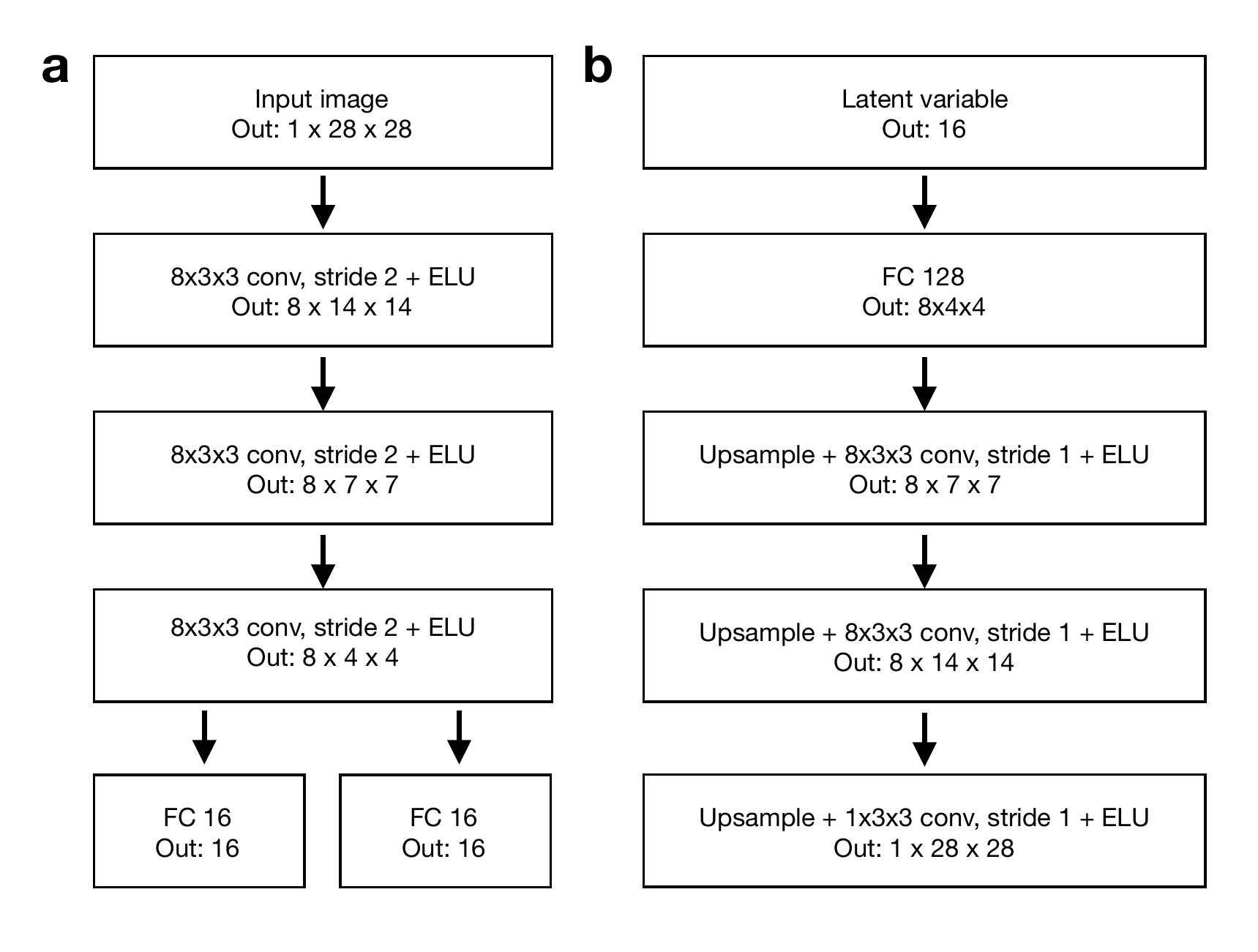}
\caption{\textbf{
Neural network architecture used in the MNIST experiments.}
(a) Encoder architecture;
(b) Decoder architecture.
The same encoder/decoder architectures are used for both VAE and \gppvae.
For CVAE, we still use the same architecture but we provide view representations (rotation angles) as inputs to both the encoder and the decoder.
For the encoder, we provide rotation angles to the first layer (by adding additional channels to the input image) and again before the dense layer (by stacking them to the vector produced by the last convolution).
Similarly, for the decoder, we provide rotation angles as inputs to the first layer (by stacking them with the 16-dimensional latent variable) and again before the first convolution layer (by adding them as additional channels to the 8x8x4 tensor fed to the first convolution layer).
For a fair comparison with \gppvae, we account for periodicity in the view representation by providing both the angle sine and cosine, i.e. $\B{w}=[\text{sin}\phi, \text{cos}\phi]$, where $\phi$ is the rotation angle.
}
\end{centering}
\end{figure}

\begin{figure}[h]
\begin{centering}
\includegraphics[width=0.6 \textwidth]{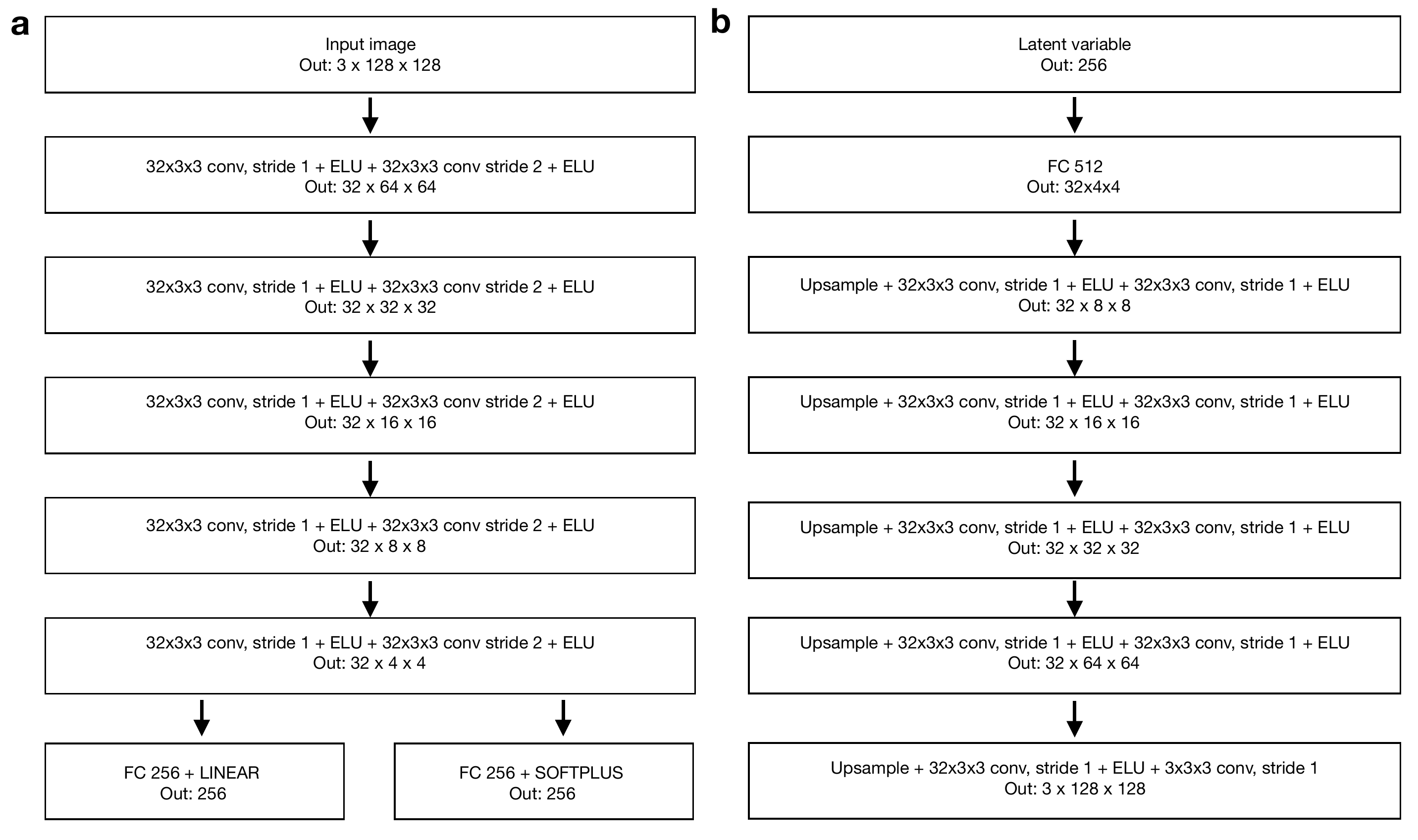}
\caption{\textbf{
Neural network architecture used in the face dataset experiments.}
(a) Encoder architecture;
(b) Decoder architecture.
The same encoder/decoder architectures are used for both VAE and \gppvae.
For CVAE, we still use the same architecture but we provide view information as one hot encoding of the 9 different poses to both the encoder and the decoder networks.
For the encoder, we provide view representations to the first layer (by adding additional channels to the input image) and again before the dense layer (by stacking them to the vector produced by the last convolution).
Similarly, for the decoder, we provide view representations as inputs to the first layer (by stacking them with the 16-dimensional latent variable) and again before the first convolution layer (by adding them as additional channels to the 8x8x4 tensor fed to the first convolution layer).
}
\end{centering}
\end{figure}

\begin{figure}[h]
\begin{centering}
\includegraphics[width=0.7 \textwidth]{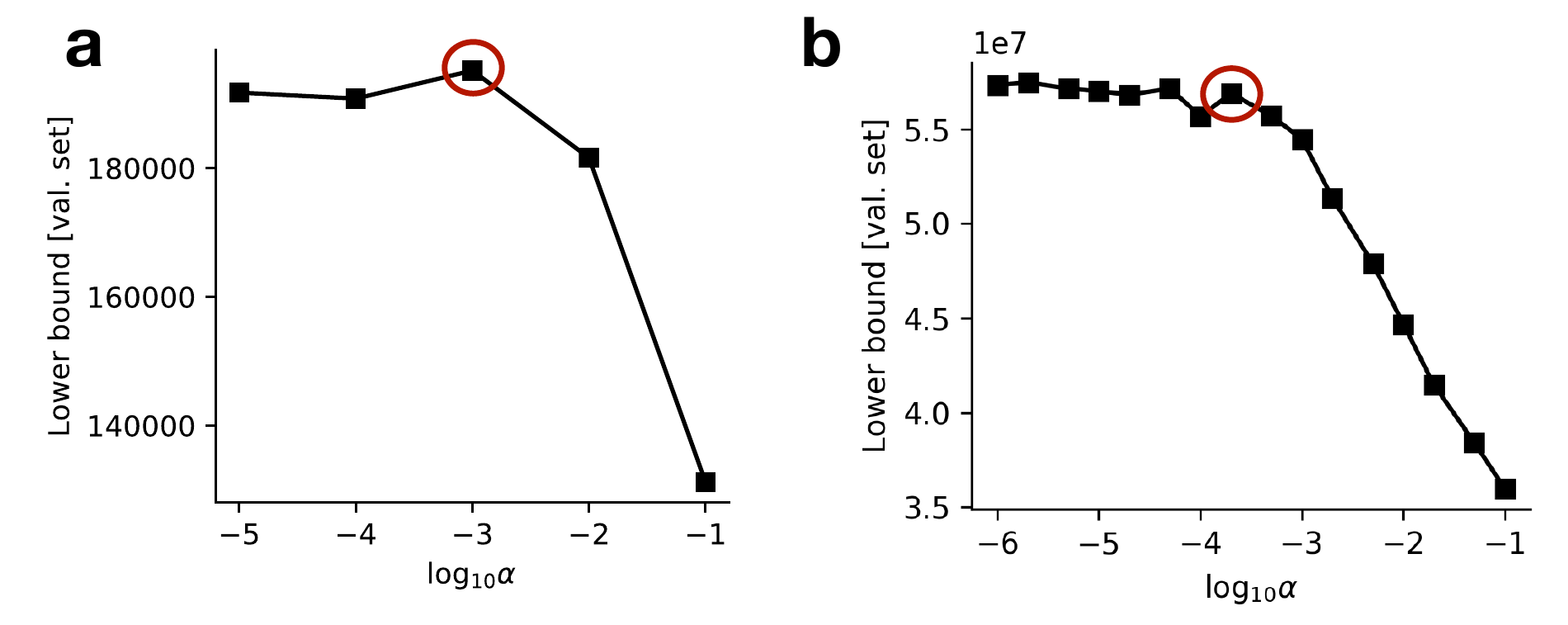}
\caption{\textbf{
Selection of the trade-off parameter between fit-to-data and latent-space model.}
The trade-off parameter was selected as to maximize VAE ELBO.
We here show the VAE validation lower bound as a function of the trade-off parameter for the MNIST (a) and the face dataset (b).
}
\end{centering}
\end{figure}

\begin{figure}[h]
\begin{centering}
\includegraphics[width=0.7 \textwidth]{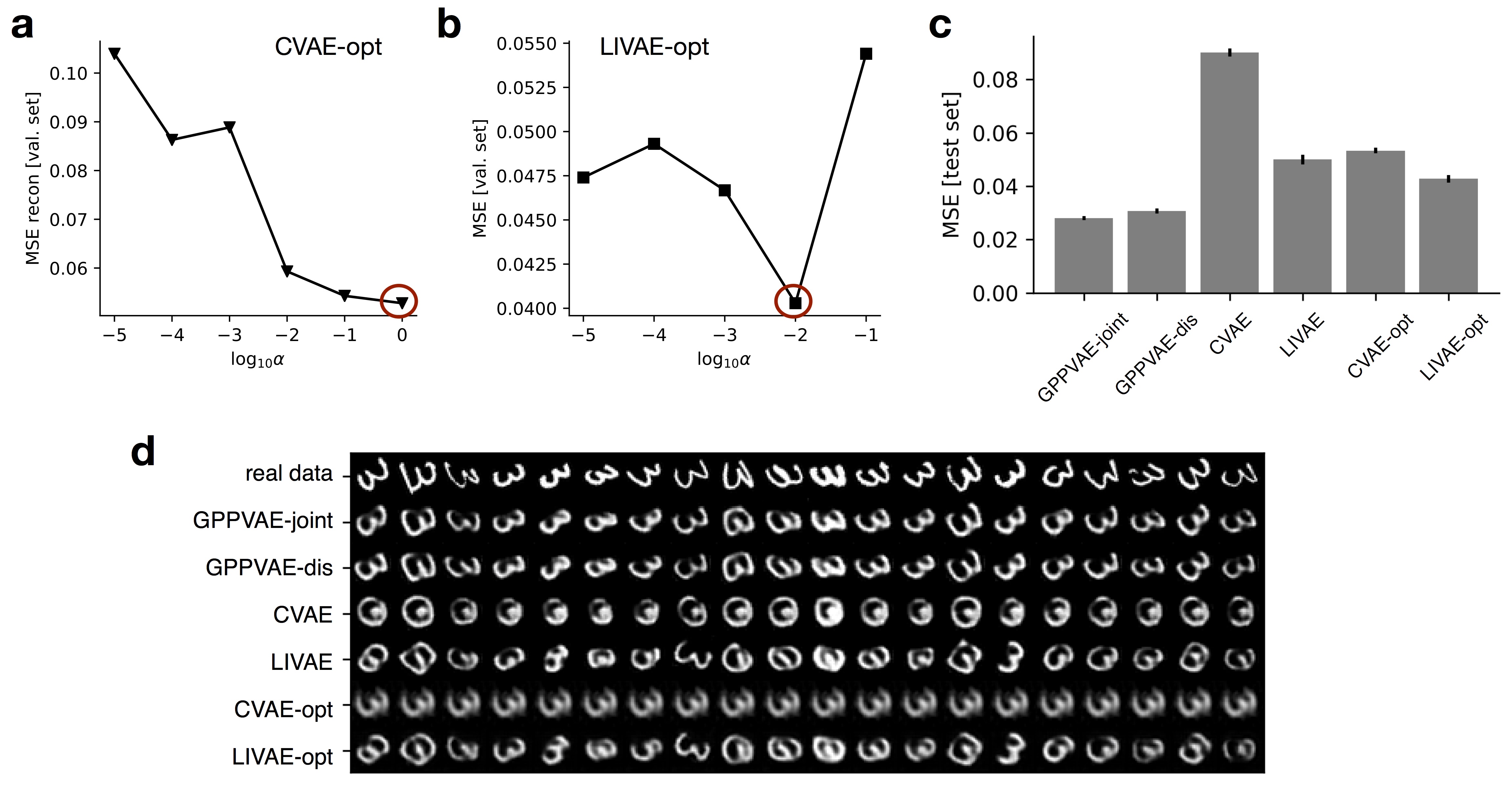}
\caption{
For CVAE and LIVAE, we also considered the alternative strategy of selecting the value of the trade-off parameter $\lambda$ that maximizes prediction performance on the validation set.
We refer to these additional methods as CVAE-opt and LIVAE-opt.
We here show the selection of $\lambda$ for CVAE-opt (a) and LIVAE-opt (b) and their performance (c-d) for the experiments in MNIST.
}
\end{centering}
\end{figure}

\begin{figure}[h]
\begin{centering}
\includegraphics[width=0.9 \textwidth]{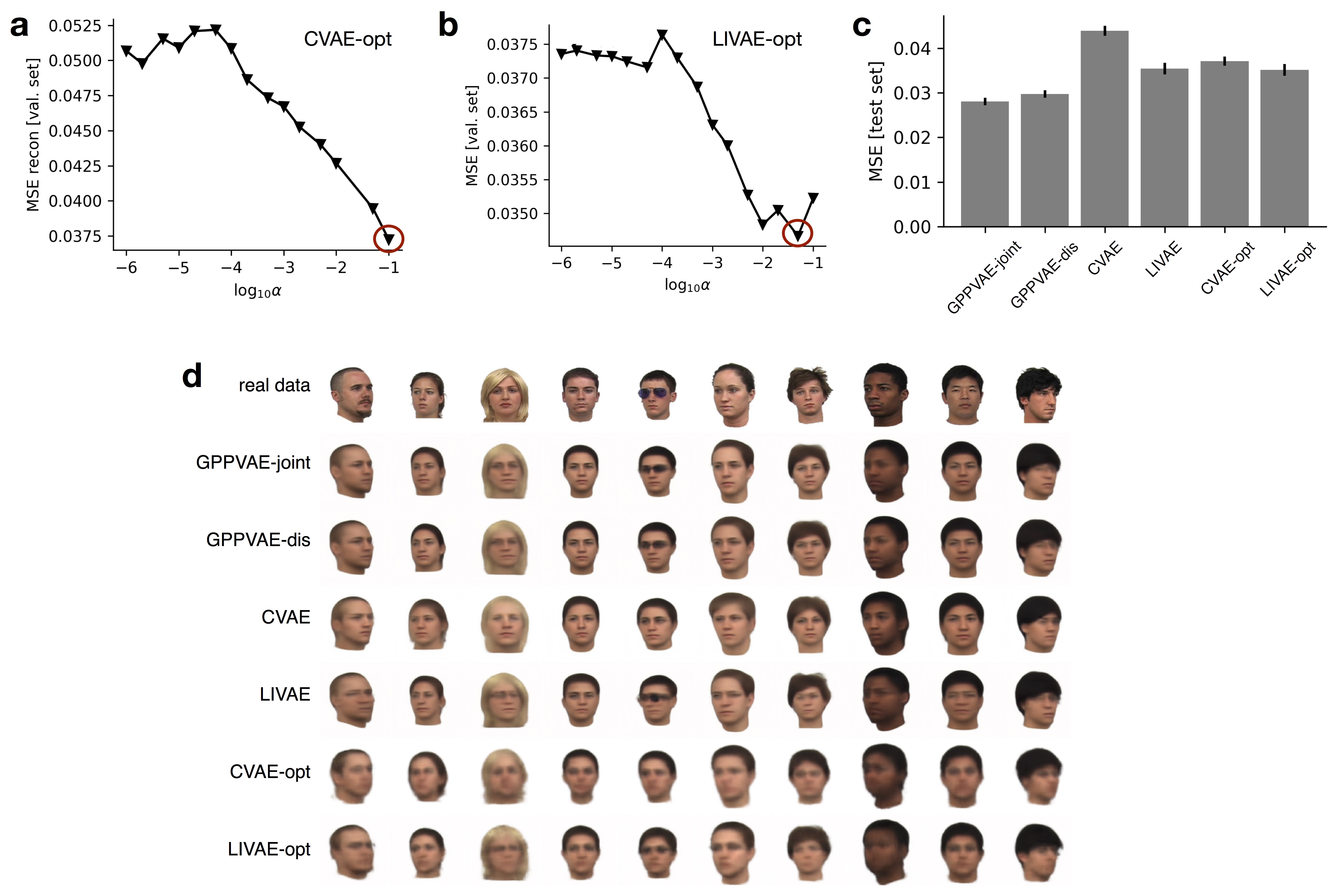}
\caption{
Analogous to Figure S4 but for the face data experiments.
}
\end{centering}
\end{figure}

\newpage

\begin{figure}[h]
\begin{centering}
\includegraphics[width=0.7 \textwidth]{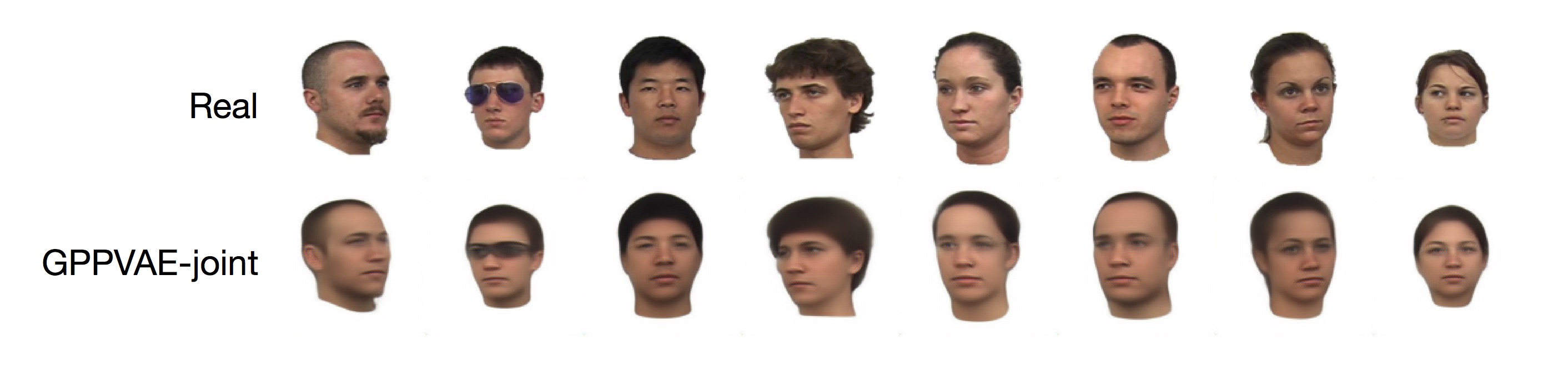}
\caption{
Out-of-sample predictions obtained from \gppvae-joint when considering the perceptual loss introduced in~\cite{hou2017deep}.
}
\end{centering}
\end{figure}

\end{appendices}


\end{document}